\documentclass[twoside]{article}

\usepackage[accepted]{aistats2016}
\usepackage{times}
\usepackage[utf8]{inputenc} 
\usepackage{amsmath}
\usepackage{amssymb}
\usepackage{dsfont}
\usepackage{caption} 
\usepackage{float} 
\usepackage{graphicx}

\newtheorem{remark}{Remark}
\newtheorem{algorithm}{Algorithm}

\newcommand{\ie}{\emph{i.e.}{}}

\newcommand\iid{\ensuremath{\mathit{i.i.d.}}\ }

\def\mb{\mathbf}
\def\point{\,\cdot\,}

\begin{document}

%
\runningtitle{Sparse Representation of Multivariate Extremes for Anomaly Detection}

%

\twocolumn[

\aistatstitle{Sparse Representation of Multivariate Extremes\\ with Applications to Anomaly Ranking}

\aistatsauthor{ Nicolas Goix \And Anne Sabourin \And Stéphan Clémençon}
~\\
\aistatsaddress{ LTCI, CNRS, Télécom ParisTech, \\ Université Paris-Saclay \And LTCI, CNRS, Télécom ParisTech, \\ Université Paris-Saclay \And LTCI, CNRS, Télécom ParisTech, \\ Université Paris-Saclay}
 ]

\begin{abstract}
 Extremes play a special role in Anomaly Detection. Beyond inference and simulation purposes, probabilistic tools borrowed from Extreme Value Theory (EVT), such as the \textit{angular measure}, can also be used to design novel statistical learning methods for Anomaly Detection/ranking. This paper proposes a new algorithm based on multivariate EVT to learn how to rank observations in a high dimensional space with respect to their degree of `abnormality'. The procedure relies on an original dimension-reduction technique in the extreme domain that possibly produces a sparse representation of multivariate extremes and allows to gain insight into the dependence structure thereof, escaping the curse of dimensionality. The representation output by the unsupervised methodology we propose here can be combined with any Anomaly Detection technique tailored to non-extreme data. As it performs linearly with the dimension and almost linearly in the data (in $O(d n \log n)$), it fits to large scale problems.
The approach in this paper is novel in that EVT has never been used in its multivariate version in the field of Anomaly Detection.
Illustrative experimental results provide strong empirical evidence of the relevance of our approach.
\end{abstract}

\section{Introduction}
\label{sec:intro}

 In an unsupervised
framework, where the dataset consists of a large number of normal
data with a smaller unknown number of anomalies, the `extreme' 
observations are more likely to be anomalies than
the others. In a supervised or semi-supervised framework, when a
dataset made of observations known to be normal is 
available, the most
extreme points delimit the outlying regions of the normal instances. 
In both cases, extreme data
 are often in a boundary region between normal and
abnormal regions and deserve special treatment. 

This angle has been intensively exploited in the one-dimensional
setting (\cite{Roberts99}, \cite{Roberts2000}, \cite{Clifton2011},
\cite{Clifton2008}, \cite{Lee2008}), where measurements are considered
as `abnormal' when they are remote from central measures such as the
mean or the median. Anomaly Detection (AD) then relies on tail analysis of
the variable of interest and naturally involves Extreme Value Theory
(EVT). Indeed, the latter builds parametric representations for the
tail of univariate distributions. In contrast, to the best of our
knowledge, \textit{multivariate} EVT has not been the subject of much
attention in the field of AD. Until now, the multivariate setup has been
treated using univariate 
extreme value statistics, to be handled with 
univariate EVT. A simple explanation is that
multivariate EVT models do not scale well with dimension: dimensionality creates difficulties for both model computation and assessment, jeopardizing machine-learning applications. In the
present paper we fill this gap by proposing a statistical method which is able to
learn a sparse `normal profile' of multivariate extremes in relation
with their (supposedly unknown) dependence structure, and, as such,
may be employed as an extension of any AD algorithm.

Since extreme observations typically constitute 
 few percents of the data, a classical AD 
algorithm would tend to classify them 
as abnormal:  it is not worth the risk (in terms of ROC curve for instance)
to try to be more precise in such low probability regions without
adapted tools. Thus, new observations 
 outside
the observed support or close to its boundary 
(larger than the largest observations) are
most often predicted
as abnormal. However, in many 
applications (\textit{e.g.} aircraft predictive maintenance), false
positives (\ie~false alarms) are very expensive, so that increasing
precision in the extremal regions is of major interest. 
 In such a context, learning the structure of extremes allows to build a `normal profile'
to be confronted with new extremal data.

In a multivariate `Peaks-over-threshold' setting, well-documented in the literature (see Chapter 9 in \cite{BGTS04} and the references therein), one observes
realizations of a $d$-dimensional r.v. $\mb X = (X_1 ,...,
X_d)$ and wants to learn the conditional distribution of excesses,
$\left[~ \mb X ~|~ \|\mb X\|_{\infty} \ge \mb u ~ \right]$ with $\|\mb X\|_{\infty}=\max_{1\leq i\leq d}\vert X_i\vert$ (notice incidentally that the present analysis could be extended to any other norm on $\mathbb{R}^d$), above some large
threshold $\mb u$. The dependence structure of such excesses is
described via the distribution of the ‘directions’ formed by the most
extreme observations - the so-called \textit{angular probability measure},
which has no natural parametric representation, 
which makes inference more complex when $d$ is large.
However, in a wide range of applications, one may expect the occurrence of two phenomena:
\noindent
\textbf{1-} Only a `small' number of groups of components may be
concomitantly extreme 
(relatively to the total number of groups~$2^d$).
\noindent
\textbf{2-} Each of these groups contains a reduced number of
coordinates (\textit{w.r.t.}. the dimension $d$). 
The main purpose of this paper is to propose a method for the statistical recovery of such
subsets, so as to reduce the dimension of the problem and thus to learn
a sparse representation of extreme -- not abnormal -- observations. 
 In the case where
hypothesis \textbf{2-} is not fulfilled, such a sparse `normal profile'
can still be learned, but it then looses the low dimensional property. 

In an unsupervised setup - namely
when data include unlabeled anomalies - 
one runs the risk of fitting the `normal profile' on 
abnormal observations. 
It is therefore essential to control the complexity of the output,
especially in a multivariate setting where EVT does not impose any
parametric form to the dependence structure. 
The method developed in this paper hence
involves a non-parametric but relatively coarse estimation 
scheme, which aims at identifying low dimensional subspaces 
 supporting extreme data. 
As a consequence, this method is robust to
outliers and also applies when the training dataset contains a (small)
proportion of anomalies.

Most of classical AD algorithms 
provide more than a predictive label, abnormal vs. normal. They return
a real valued function, inducing a preorder/ranking on the input
space. Indeed, when confronted with massive data, being able to rank
observations according to their supposed degree of abnormality may
significantly improve operational processes and allow for a
prioritization of actions to be taken, especially in situations where
human expertise required to check each observation is time-consuming
(\textit{e.g.} fleet management). Choosing a threshold for the
ranking function yields a 
decision function delimiting normal regions from abnormal ones. The
algorithm proposed 
in this paper deals with this problem of \textit{anomaly ranking} and
provides a ranking function (also termed a \textit{scoring function})
for extreme observations. 
 This method is complementary to other AD algorithms in the sense that 
a standard AD scoring function may be learned using the `non extreme' (below
threshold) observations of a dataset, while `extreme' (above
threshold) data are used to learn an extreme scoring
function. 
Experiments on classical AD datasets show a significant performance 
improvement 
in terms of precision-recall curve, while preserving undiluted ROC curves.
As expected, the \textit{precision} of the standard AD algorithm is
improved in extremal regions, since the algorithm `takes the risk'
not to consider systematically as abnormal the extremal regions, and
to adapt to
 the specific structure of extremes instead. 
These improvements
may typically be useful in applications where the cost of false positives (\ie~false alarms) is very expensive.

The structure of the paper is as follows. The algorithm is presented
in Section~\ref{sec:algo}. In Section~\ref{sec:framework}, the whys
and wherefores of EVT connected to the present analysis are recalled
before a rationale behind the estimation involved by the
algorithm. Experiments on both simulated and real datasets are
performed respectively in Section~\ref{sec:experiments-simulated} and
\ref{sec:experiments-real}.

\section{A dependence-based AD algorithm}
\label{sec:algo} 
The purpose of the algorithm presented below is to rank multivariate
extreme observations, based on their dependence structure. The
present section details the algorithm and provides the heuristic of the
mechanism at work,  which can be understood without knowledge of EVT. A
theoretical justification which does rely on EVT is given in Section~\ref{sec:framework}. The underlying assumption is that an
observation is potentially abnormal if its `direction' (after a standardization of each marginal) is special
regarding to the other extreme observations. In other words, if it
does not belong to the (sparse) support of
extremes. 
Based on this
intuition, a scoring function is built to compare the degree of
abnormality of extreme observations. 

Let $\mb X_1,.\ldots,\mb X_n$ \iid random variables in $\mathbb{R}^d$ with joint
(\emph{resp.} marginal) distribution $F$ (\emph{resp.} $F_j$, $j =
1,\ldots,d$). Marginal standardization is a natural first step when
studying the dependence structure in a multivariate setting. The
choice of standard Pareto margins $V^j$ (with $\mathbb{P}(V^j > x ) =
1/x$, $x>0$) is convenient -- this will become clear in
Section~\ref{sec:framework}. One classical way to standardize is the
probability integral transform, $T : \mb X_i \mapsto \mb V_i = ( (1- F_j
(X_i^j))^{-1})_{1\leq j\leq d}$, $i = 1,\ldots,n$. 
Since the marginal distributions $F_j$ are unknown, we use their
empirical counterpart $\hat F_j$, where $\hat F_j (x) =
(1/n) \sum_{i=1}^n \mathds{1}_{X_i^j\le x}$. 
Denote by $\hat T$ the rank transformation
thus obtained and by $\hat{\mb V}_i = \hat T(\mb X_i)$ the corresponding rank-transformed
observations. 




Now, the goal is to measure the
`correlation' within each
subset of features $\alpha \subset \{1,...,d\}$ at extreme levels (each $\alpha$ corresponding to a sub-cone of the positive orthant), that
is, the likelihood to observe a large $\mathbf{\hat V}$ which verifies the following condition: $\hat V^j$ is `large' for all
$j\in\alpha$, while the other $\hat V^j$'s ($j \notin \alpha$) are `small'.
Formally, one may 
associate to each such $\alpha$ a coefficient
reflecting the degree of dependence between the
features $\alpha$ at extreme levels. 
In relation to Section~\ref{sec:framework}, the appropriate way to
give a meaning to `large' (\emph{resp.} `small') among extremes is in
`radial' and `directional' terms, that is, $\| \mathbf{\hat V}\|>r$ (for some
high radial threshold $r$), and $\hat V^j/\|\mathbf{\hat V}\| >\epsilon$
(\emph{resp.} $ \le \epsilon$) for some small directional tolerance
parameter 
$\epsilon>0$. Note that $ \mathbf{\hat V}/\|\mathbf{\hat V}\|$ has unit norm and can be viewed as
the pseudo-angle of the transformed data $\mathbf{\hat V}$. 
Introduce the truncated $\epsilon$-cones (see Fig.~\ref{2Dcones}):
\begin{align}
 \label{eq:epsilonCone}
 ~\mathcal{C}_\alpha^\epsilon~=~&\big\{\mb v \ge
 0,~\|\mb v\|_\infty \ge 1,~ v_i > \epsilon \|\mb v\|_\infty ~\text{ for } i
 \in \alpha,\\
\nonumber &~~~~~~~~~~~~~~~~~~~~~~~~~~~~~~~~~~v_i \le \epsilon\|\mb v\|_\infty ~\text{ for } i \notin \alpha
~~ \big\}~,
\end{align}
which defines a
partition of $\mathbb{R}_+^{d}\setminus [0,1]^d$ for each fixed $\epsilon\ge 0$. 
%
%
This leads to  coefficients 
\begin{align}
\label{mu_n}
\mu_n^{\alpha, \epsilon} = (n/k) \mathbb{\hat P}_n \left (
  (n/k) \mathcal{C}_\alpha^\epsilon \right), 
\end{align}
\noindent
where $\mathbb{\hat P}_n(.)=(1/n)\sum_{i=1}^n\delta_{\hat{V}_i}(.)$ is the empirical probability distribution of the rank-transformed data and $k =
k(n) \to \infty$ s.t. $k = o(n)$ as $n \to
\infty$. 
The ratio $n/k$ plays the role of a large radial
threshold $r$. From our standardization choice, counting points in
$(n/k)\,\mathcal{C}_{\alpha}^{\epsilon}$ boils down to
selecting, for each feature $j\le d$,   the `$k$ largest values' $X_i^j$
over the $n$ observations, whence the normalizing factor $\frac{n}{k}$. 
In an Anomaly Detection framework, the degree of `abnormality' of new observation $\mb x$ such that
$\hat T(\mb x) \in \mathcal{C}_\alpha^\epsilon$ 
should be related both to $\mu_n^{\alpha, \epsilon}$ and the uniform
norm $\|\hat T(\mb x)\|_\infty$ (angular and radial
components). As a matter of fact, in the transformed space - namely the
space of the $\hat V_i$'s - the asymptotic mass decreases as the
inverse of the norm, see~(\ref{mu-phi}).
Consider the  `\textit{directional tail region}' induced by $\mb x$, 
$A_{\mb x} =  \{\mb y  :
T(\mb y) \in \mathcal{C}_\alpha^\epsilon\,,\;\|T(\mb y)\|_\infty \ge \|
T(\mb x)\|_\infty\}$ where $\mb x \in \mathcal{C}_\alpha^\epsilon$.
 Then, if  $\|T(\mb x)\|_\infty$ is large enough, we
shall see (as \emph{e.g.} in  \eqref{eq:understandSn}) that 
$  \mathbb{P}( \mb X \in A_{\mb x} ) \simeq  \|\hat T(\mb x) \|_\infty^{-1} \mu_n^{\alpha,\epsilon} $.
%
%
 This yields the scoring
function $s_n(\mb x)$
(\ref{def:scoring}), which is thus an empirical version of
$\mathbb{P}(\mb X\in A_{\mb x})$: the smaller $s_n(\mb x)$, the more abnormal the point $\mb x$ should be considered.

This heuristic yields the following algorithm, referred to as the {\it Detecting Anomaly with Multivariate EXtremes} algorithm (DAMEX in abbreviated form).
The complexity is in $O( dn\log n + dn) = O(dn\log n)$, where the
first term on the left-hand-side comes
from  computing the $\hat F_j(X_i^j)$ (Step 1) by sorting  the data
(\emph{e.g.} merge sort). The second one comes from Step 2. 
\noindent
\begin{remark}({\sc Interpretation of the Parameters})
\label{rk_param_interpretation}
In view of (\ref{eq:epsilonCone}) and (\ref{mu_n}), $n/k$ is the threshold beyond which the data are considered as extreme. 
A general heuristic in multivariate extremes is that $k$ is proportional to the number of data considered as extreme. $\epsilon$ is the tolerance parameter \textit{w.r.t.}. the non-asymptotic nature of data. The smaller $k$, the smaller $\epsilon$ shall be chosen.
\end{remark}
\begin{remark}({\sc Choice of Parameters})
\label{rk_param_choice}
There is no simple manner to choose the parameters $(\epsilon,~ k ,~ \mu_{\min})$, as there is no simple way to determine how fast is the convergence to the (asymptotic) extreme behavior --namely how far in the tail appears the asymptotic dependence structure.
In a supervised or semi-supervised framework (or if a small labeled dataset is available) these three parameters should be chosen by cross-validation.
In the unsupervised situation, a classical heuristic (\cite{Coles2001}) is to choose $(k, \epsilon)$ in a stability region of the algorithm's output: the largest $k$ (resp. the larger $\epsilon$) such that when decreased, the dependence structure remains stable. Here, `stable' means that the sub-cones with positive mass do not change much when the parameter varies in such region.
 This amounts to selecting the maximal number of data to be extreme, constrained to observing the stability induced by the asymptotic behavior. 
Alternatively, cross-validation can still be used in the unsupervised framework, considering one-class criteria such as the Mass-Volume curve or the Excess-Mass curve (\cite{AISTAT15, CLEM13}), which play the same role as the ROC curve when no label is available. As estimating such criteria involve some volume estimation, a stepwise approximation (on hypercubes, whose volume is known) of the scoring function should be used in large dimension.
\end{remark}
\begin{remark} ({\sc Dimension Reduction})
If the extreme dependence structure is low dimensional, namely
concentrated on low dimensional cones $\mathcal{C}_\alpha$ -- or in other terms if only a
limited number of margins can be large together -- then most of the
$\hat V_i$'s will be concentrated on $\mathcal{C}_\alpha^\epsilon$'s
such that  $|\alpha|$ (the dimension of the cone $\mathcal{C}_\alpha$)
is small; then the
representation of the dependence structure
 in (\ref{phi_n}) is both sparse and low dimensional.
\end{remark}

\begin{center}
\fbox{
\begin{minipage}{0.95\linewidth}
\begin{algorithm} (DAMEX)
\label{DAMEX-algo}\\
{\bf Input:} parameters $\epsilon>0$,~~ $k = k(n)$,~~ $\mu_{\min}\geq 0$.
\begin{enumerate}
\item Standardize \emph{via} marginal rank-transformation: $\hat V_i:= \big (1/(1- \hat F_j (X_i^j))\big)_{j=1,\ldots,d}$~. 
\item Assign to each $\hat V_i$ the cone $\mathcal{C}_\alpha^\epsilon$
  it belongs to.  
\item Compute $\mu_n^{\alpha, \epsilon}$ from (\ref{mu_n})
   $\rightarrow$ yields: (small number of) cones with non-zero mass
\item Set to $0$ the $\mu_n^{\alpha,\epsilon}$ below some small
  threshold $\mu_{\min}\ge 0$ to eliminate cones with negligible mass
   $\rightarrow$ yields: (sparse) representation of the dependence
  structure 
 \begin{align}
 \label{phi_n}
(\mu_n^{\alpha,\epsilon})_{\alpha\subset\{1,\ldots, d\}, \mu_n^{\alpha,\epsilon}>\mu_{\min}}
 \end{align}
\end{enumerate}
{\bf Output:} Compute the scoring function given by 
\begin{align}
\label{def:scoring}
s_n(\mb x):= (1/\|\hat T(\mb x)\|_\infty)
\sum_{\alpha }
\mu_n^{\alpha, \epsilon} \mathds{1}_{\hat T(\mb x) \in \mathcal{C}_\alpha^\epsilon}.
\end{align}
\end{algorithm}
\end{minipage}
}
\end{center}

The next section provides a theoretical ground for Algorithm
\ref{DAMEX-algo}. 
As shall be shown below, it  amounts to
learning  the dependence structure of extremes (in particular, its support).
The dependence parameter $\mu_n^{\alpha,\epsilon}$ 
actually coincides with a
(voluntarily $\epsilon$-biased) natural estimator of $\mu(\mathcal{C}_\alpha)$, where $\mu$
is  a `true' measure of the extremal dependence and
$\mathcal{C}_\alpha$ is the truncated cone obtained with $\epsilon =
0$ (Fig.~\ref{fig:3Dcones}),  
\noindent
\begin{align}
\label{cone}
&\mathcal{C}_\alpha = \big\{\mb x \ge 0 : ~\|\mb x\|_\infty \ge 1,~ x_i > 0 \text{ for } i \in \alpha,\\
\nonumber &~~~~~~~~~~~~~~~~~~~~~~~~~~~~~~~~~~~~~~~~~~~~~~ x_i = 0 ~\text{ for } i \notin \alpha ~~~\big\}.
\end{align}
\section{Theoretical framework}
\label{sec:framework}
\subsection{Probabilistic background}

\paragraph{Univariate and multivariate EVT}
Extreme Value Theory (\textsc{EVT}) develops models for learning  the
unusual rather than the usual. These  models are widely used in fields
involving risk management like finance, insurance, telecommunication
or environmental sciences. 
One major application of \textsc{EVT} is to provide a reasonable
assessment of the probability of occurrence of rare events. 
A useful setting to understand the use  of \textsc{EVT} is that  of
risk monitoring. 
A typical  quantity of interest in the univariate case is  the $(1-p)^{th}$
quantile of the distribution  $F$ of a random
variable $X$,  for a given exceedance probability $p$, that is
$x_p = \inf\{x \in \mathbb{R},~ \mathbb{P}(X > x) \le p\}$. For
moderate values of $p$, a natural empirical estimate is  $x_{p,n} = \inf\{x \in
\mathbb{R},~ 1/n \sum_{i=1}^n \mathds{1}_{X_i > x}\le p\}$.
However,  if
$p$ is very small, the finite  sample $X_1,
\ldots, X_n$  contains insufficient information and $x_{p,n}$ becomes 
irrelevant. 
That is where \textsc{EVT} comes into play  by providing
parametric estimates of large
quantiles: 
in this case,  \textsc{EVT} essentially consists in modeling
the distribution of the maxima (\emph{resp.} the upper tail) as a Generalized
Extreme Value (GEV) distribution, namely an element of the Gumbel, Fréchet
or Weibull parametric families (\emph{resp.} by a generalized Pareto distribution).
Whereas statistical inference often involves sample means and the
central limit
theorem, 
\textsc{EVT} handles phenomena whose behavior is 
not ruled by an `averaging effect'. The focus is on large quantiles 
rather than the mean. The primal -- and not too stringent -- assumption is the existence of two
sequences $\{a_n, n \ge 1\}$ and $\{b_n, n \ge 1\}$, the $a_n$'s being
positive, and a non-degenerate cumulative distribution function (\emph{c.d.f.}) $G$ such that
\begin{align}
\label{intro:assumption1}
\lim_{n \to \infty} n ~\mathbb{P}\left( \frac{X - b_n}{a_n} ~\ge~ x \right) = -\log G(x) 
\end{align}
\noindent
for all continuity points $x \in \mathbb{R}$ of $G$.
If assumption~\eqref{intro:assumption1} is fulfilled -- it is the case for most  textbook
distributions --  $F$ is said to be in the \textit{domain of attraction} of $G$, denoted $F \in DA(G)$. 
The tail behavior of $F$
is then essentially characterized by $G$, which is proved to 
belong to the parametric family of GEV distributions, namely to be -- up
to  rescaling -- of the type $G(x) = \exp(-(1 + \gamma
x)^{-1/\gamma})$ for $1 + \gamma x > 0$, $\gamma \in \mathbb{R}$,
setting by convention $(1 + \gamma x)^{-1/\gamma} = e^{-x}$ for
$\gamma = 0$. The sign of $\gamma$ controls the shape of the tail and
various estimators of the rescaling sequence as well as $\gamma$ have
been studied in great detail,
see \emph{e.g.}  
\cite{Hill1975}, \cite{Smith1987}, \cite{BVT1996}. 

The multivariate analogue of the assumption~\eqref{intro:assumption1} concerns, again,
the convergence of the tail probabilities, namely, 
\begin{align}
\label{intro:assumption2}
&\lim_{n \to \infty} n ~\mathbb{P}\left( \frac{X^1 - b_n^1}{a_n^1} ~\ge~ x_1  \;\text{~or~} \ldots \text{~or~} \;\frac{X^d - b_n^d}{a_n^d} ~\ge~  x_d \right) \\
\nonumber &~~~~~~~~~~~~~~~~~~~~~~~~~~~~~~~~~~~~~~~~~~~~~~~~~~~~~~~~~~~~~~~~~~=~ -\log \mb G(\mathbf{x}),
\end{align}
(denoted $\mb F \in \textbf{DA} (\mb G)$) for all continuity points $\mathbf{x} \in \mathbb{R}^d$ of $\mb G$.
Here $a_n^j>0$ and $\mb G$ is a non
degenerate multivariate \emph{c.d.f.}. 
%
This implies
that the margins $G_1(x_1),\ldots,G_d(x_d)$ are univariate extreme
value distributions, namely of the type $G_j(x) = \exp(-(1 + \gamma_j
x)^{-1/\gamma_j})$. 
 Also, denoting by $F_1,\ldots,F_d$ the
marginal
distributions of $\mb F$, assumption (\ref{intro:assumption2}) implies
marginal convergence, $F_i \in DA(G_i)$ for $i=1,\ldots,n$.
However, extending the theory and estimation methods from the
univariate case is far from obvious, since  the 
 dependence structure of the joint distribution $\mb G$ comes into play
 and has no exact finite-dimensional parametrization.

\paragraph{Standardization and Angular measure}
 To understand the form  of the limit $\mb G$ and dispose of the
 unknown sequences $(a_n^j,  b_n^j)$, 
 it is most convenient to work with marginally standardized variables,
 $V^j:=\frac{1}{1-F_j(X^j)}$ and $\mathbf{V}=(V^1,\ldots,V^d)$, as
 introduced in Section~\ref{sec:algo}.  In fact (see
 \cite{Resnick1987}, Proposition 5.10), the multivariate tail
 convergence assumption in
 (\ref{intro:assumption2}) is equivalent to marginal convergences $F_j
 \in \text{DA}(G_j)$ as in
 (\ref{intro:assumption1}), 
 together with regular variation of the tail of 
  $\mathbf{V}$, \ie~there exists a limit measure
 $\mu$ on $\mb E = [0,\infty]^d\setminus\{0\}$, such
 that 
\begin{align}
\label{intro:regvar}
n~ \mathbb{P}\left( \frac{V^1 }{n} ~\ge~ v_1 \text{~or~} \cdots
   \text{~or~} \frac{V^d }{n} ~\ge~ v_d \right) \xrightarrow[n\to\infty]{}\mu[0,\mb v]^c
\end{align}
\noindent (where $[0,\mathbf{v}]=[0,v_1]\times \ldots \times
[0,v_d]$). Thus, the variable $\mb V$ satisfies
 (\ref{intro:assumption2}) with $\mb a_n = (n,\ldots,n)$, $\mb b_n = (0,\ldots,0)$.
The so-called \emph{exponent measure} $\mu$  has the 
homogeneity property: $\mu(t\point) =
t^{-1}\mu(\point)$. To wit, $\mu$ is, up to a normalizing factor,
the asymptotic distribution of $\mb V$ on extreme regions, that is, for
large $t$ and any fixed region $A$ bounded away from $0$, we have 
\begin{align}
\label{eq:approx_mu}
t\mathbb{P}(\mb V\in t A)\simeq \mu(A).
\end{align}

Notice that  the  limit joint \emph{c.d.f.} $G$ 
can be retrieved  from $\mu$ and the margins of $G$, \emph{via}  
 $ - \log G(\mathbf{x})= \mu\left[ \mb 0, \left(\frac{-1}{\log G_1(x_1)},
     \dots ,\frac{-1}{\log G_d(x_d)}\right)\right]^c$.  %
The choice of a marginal standardization  to handle $V^j$'s
variables is somewhat arbitrary and alternative standardizations lead
to alternative limits. 

Using the homogeneity property $\mu(t\point) =
t^{-1}\mu(\point)$, it can be shown (see \emph{e.g.} \cite{dR1977})
that in pseudo-polar coordinates, the  radial and angular
components of $\mu$ are independent: 
For $(v_1,...,v_d) \in \mathbf{E}$, let 
\begin{align*}
&R(\mb v):= \|\mb v\|_\infty ~=~ \max_{i=1}^d v_i \\
\text {~~~~and~~~~} &\Theta (\mb v) := \left( \frac{v_1}{R(\mb v)},..., \frac{v_d}{R(\mb v)} \right) \in S^\infty_{d-1}
\end{align*}
where $S^\infty_{d-1}$ is the unit sphere in $\mathbb{R}^d$ for the
infinity norm.  Define the \emph{angular measure} $\Phi$ (also called
\emph{spectral measure}) by $\Phi(B)= \mu (\{\mb v : R(\mb v)>1 , \Theta(\mb v)
\in B \})$, $B\in  S^\infty_{d-1}$.
Then, by homogeneity,  
\begin{align}
\label{mu-phi}
\mu( R>r, \Theta \in B ) = r^{-1} \Phi (B)~. 
\end{align}
In a nutshell,  there
is a one-to-one correspondence between 
$\mu$ 
 and the angular measure
$\Phi$, and any one of them can be used to characterize  the 
asymptotic tail dependence of the distribution $F$. 

\begin{figure}[!ht]
\centering
\includegraphics[scale=0.15]{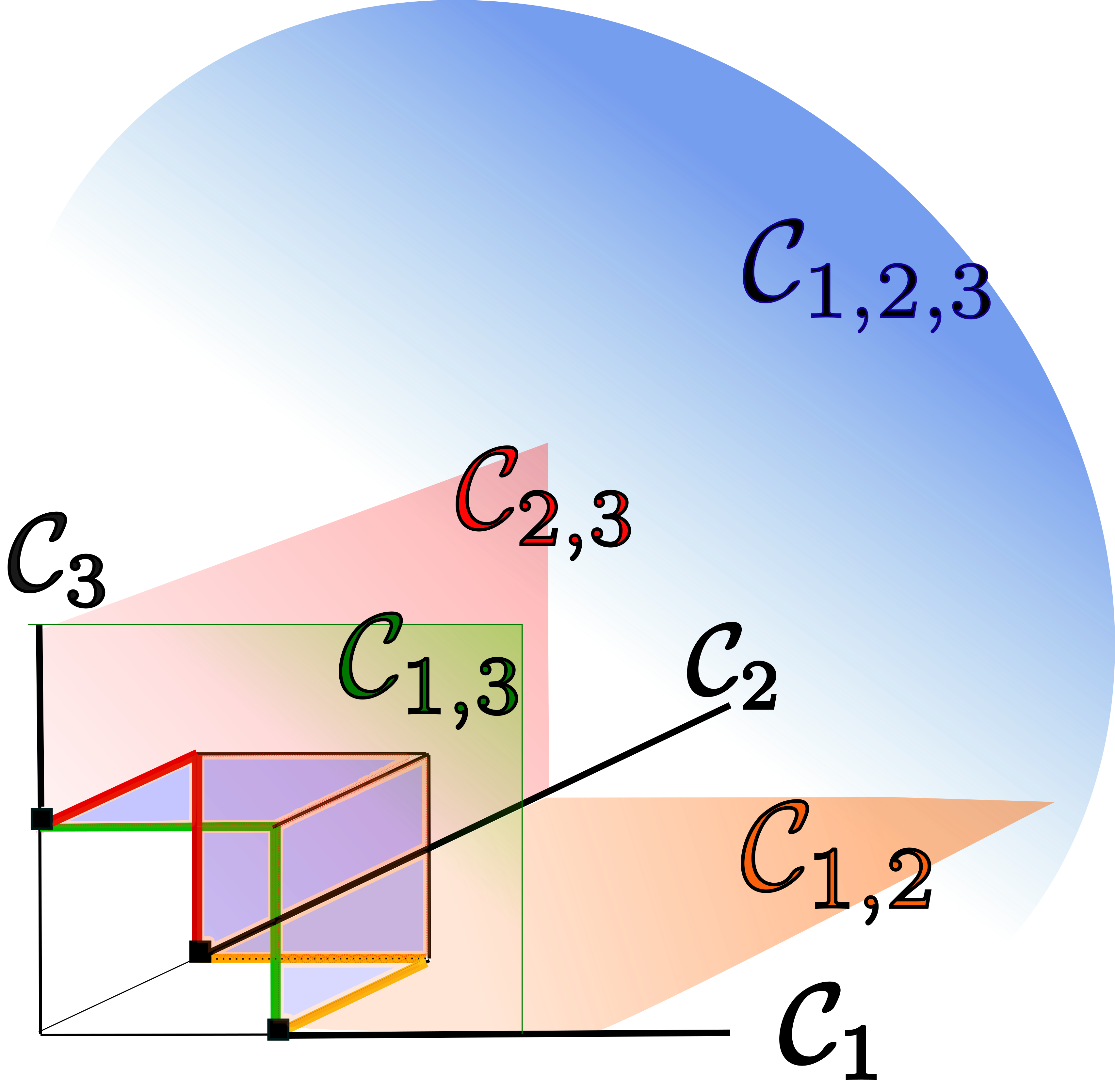}
\caption{Truncated cones in 3D}
\label{fig:3Dcones}
\end{figure}

\paragraph{Sparse support} For $\alpha$  a nonempty subset of
$\{1,\ldots ,d\}$ consider the truncated cones $\mathcal{C}_\alpha$
 defined by Eq.~(\ref{cone}) in the previous section
and illustrated in Fig.~\ref{fig:3Dcones}.
The family $\{\mathcal{C}_\alpha, \alpha\subset\{1,\ldots ,d\},
\alpha\neq\emptyset\}$   defines a partition of
$\mathbb{R}_+^{d}\setminus [0,1]^d$. 
In theory, $\mu$ may possibly allocate some mass on each cone $\mathcal{C}_\alpha$. 
A non-zero value of the cone's mass $\mu(\mathcal{C}_\alpha)$ indicates
that it is not abnormal to record simultaneously large values of the
coordinates $X^j, j\in\alpha$, together with simultaneously small
values of the complementary features $X^j, j \notin\alpha$. On the
contrary, zero mass on the cone
$\mathcal{C}_\alpha$ (\ie~, $\mu_\alpha = 0$) indicates that such
records would be abnormal. A reasonable assumption in a lot of large
dimensional use cases is that $\mu(\mathcal{C}_\alpha) = 0$ for the
vast majority 
of the $2^d -1$ cones $\mathcal{C}_\alpha$, especially for large $|\alpha|$'s.

\begin{figure}[!ht]
\centering
\includegraphics[width=0.5\linewidth]{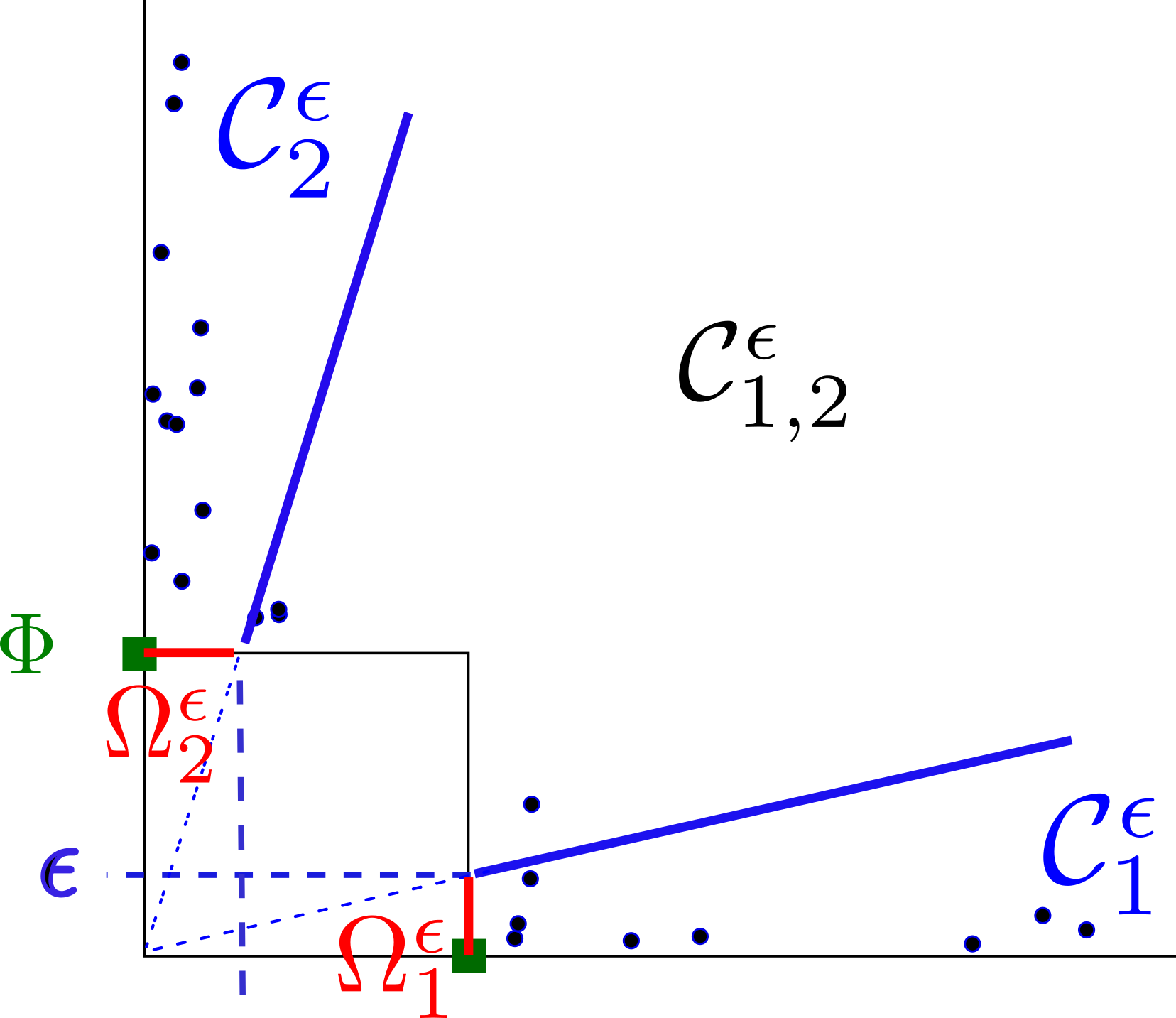}
\caption{Truncated $\epsilon$-cones in 2D}
\label{2Dcones}
\end{figure}

Equivalently, the angular measure $\Phi$ defined in~\eqref{mu-phi} decomposes as $ \Phi ~=~ \sum_{\emptyset
   \subsetneq \alpha\subset\{1,\ldots ,d\}} \Phi_\alpha $ 
corresponding to the partition
$S_{d-1}^\infty
 =\coprod_{\emptyset\neq\alpha\subset\{1,\ldots,d\}} {\Omega}_\alpha$ of the unit sphere,
 where ${\Omega}_\alpha= S_{d-1}^\infty\cap {\mathcal{C}}_\alpha$.
 Our aim is to learn the support of $\mu$ 
 or, more precisely, which cones 
 have non-zero total mass $\mu(\mathcal{C}_\alpha)=\Phi(\Omega_\alpha)$.  
The
next subsection shows that the $\mu_{n}^{\alpha,\epsilon}$'s
introduced in Algorithm~\ref{DAMEX-algo} are  empirical estimators of the
$\mu(\mathcal{C}_\alpha)$'s. 
\noindent
\subsection{Estimation of $\mu(\mathcal{C}_\alpha)$}
\label{sec:estimation}

We point out that as soon as $\alpha \neq \{1,\ldots,d\}$, the cone
$\mathcal{C}_\alpha$ is a subspace of zero Lebesgue measure.  Real data,
namely non-asymptotic data, generally do not concentrate on such sets,
so that, if we were simply counting the points $\hat V_i$ in
$\mathcal{C}_\alpha$, only the largest dimensional cone (the central
one, corresponding to $\alpha= \{1,\ldots,d\})$ would have non zero
mass. The idea is to introduce  a tolerance parameter $\epsilon$ in order to
capture the points whose projections on the unit sphere are $\epsilon$-close
to the cone  $\mathcal{C}_\alpha$, as illustrated in Fig.~\ref{2Dcones}.
This amounts to defining $\epsilon$-thickened faces on the sphere
$S_\infty^{d-1}$, 
$\Omega_\alpha^\epsilon =
\mathcal{C}_\alpha^\epsilon\cap S_\infty^{d-1}$ (the projections of the
cones defined in~\eqref{eq:epsilonCone} onto the sphere), so that  
$$ \Omega_{\alpha}^\epsilon = \big\{\mb x \in S_{d-1}^\infty , ~x_i >
\epsilon \text{ for } i\in\alpha~,~  x_i \le \epsilon \text{ for } i\notin \alpha   \big\}.$$
A natural estimator of $\mu(\mathcal{C}_\alpha)$ is thus
$\mu_n(\mathcal{C}_{\alpha}^\epsilon) = \mu_n^{\alpha,\epsilon}$, as
  defined in Section~\ref{sec:algo}, see Eq.~\eqref{mu_n} therein.
%
Thus, if $\mb x$ is a new observation,
and if $\hat T(\mb x)$ belongs to the $\epsilon$-thickened cone
$\mathcal{C}^{\epsilon}_\alpha$ defined in~\eqref{eq:epsilonCone}, the scoring function
$s_n(\mb x)$ in \eqref{def:scoring} is in fact an empirical version of the quantity 
\[\mathbb{P}(\mb X\in A_{\mb x}) := \mathbb{P}(T(\mb X)\in \mathcal{C}_\alpha, \| T(\mb X) \|_\infty >
\|T(\mb x)\|).\]
Indeed, the latter is (using \eqref{eq:approx_mu}) 
\begin{align}
  \label{eq:understandSn}
 \mathbb{P}(\mb V\in \|T(\mb x)\|_\infty \mathcal{C}_\alpha) &~~\simeq~~  \|T(\mb x)\|_\infty^{-1}\mu( \mathcal{C}_\alpha) \\
\nonumber &~~\simeq~~    \|\hat T(\mb x)\|_\infty^{-1}\mu_n^{\alpha,\epsilon} ~~=~~  s_n(\mb x) 
\end{align}

It is beyond the scope of this paper to investigate upper bounds
for $|\mu_n(\mathcal{C}_\alpha^\epsilon)-\mu(\mathcal{C}_\alpha)|$, which should be based on the decomposition:
$|\mu_n(\mathcal{C}_\alpha^\epsilon) - \mu(\mathcal{C}_\alpha)| ~~\le~~  |\mu_n- \mu|(\mathcal{C}_\alpha^\epsilon) ~~+~~|\mu(\mathcal{C}_{\alpha}^\epsilon) - \mu(\mathcal{C}_\alpha) | .$
The argument would be to investigate the first term in the right hand
size 
by approximating the sub-cones
$\mathcal{C}_{\alpha}^\epsilon$ by a Vapnik-Chervonenkis (VC) class of rectangles like in
\cite{COLT15}, where a non-asymptotic bound is stated on the
estimation of the so-called \textit{stable tail dependence
  function} (which is just another version of the exponent measure,
using a standardization to uniform margins instead of Pareto margins). 
As for the second term, since
$\mathcal{C}_\alpha^\epsilon$ and $\mathcal{C}_\alpha$ are
close up to a volume proportional to $\epsilon^{d - |\alpha|}$, their mass can be
proved close, under the condition that the density of $\mu|_{\mathcal{C}_\alpha}$
\textit{w.r.t.} Lebesgue measure of dimension $|\alpha|$  on
$\mathcal{C}_\alpha$  is bounded.


\section{Experiments on simulated data}
\label{sec:experiments-simulated}

\subsection{Simulation on 2D data}
\begin{figure}[!ht]
\centering
\includegraphics[scale=0.325]{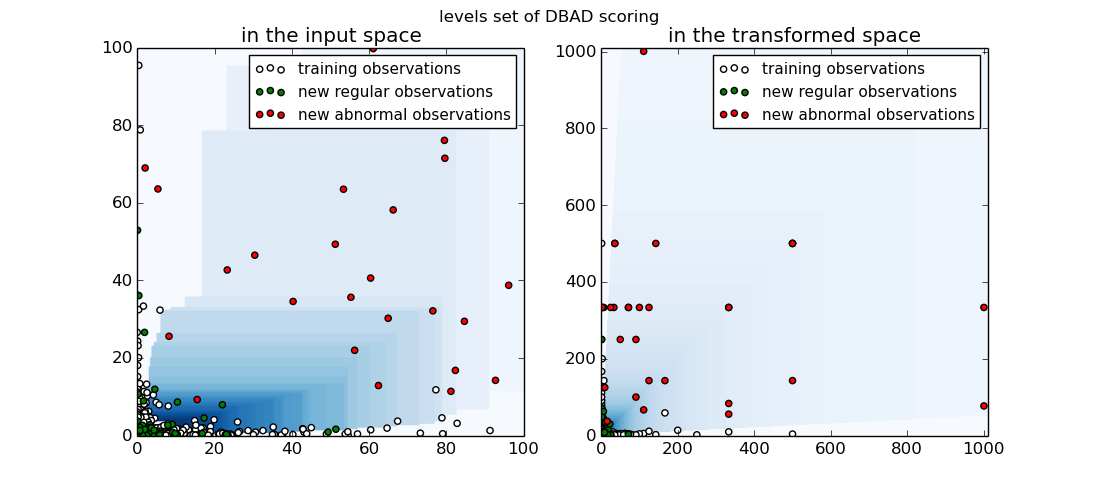}
\caption{Level sets of $s_n$ on simulated 2D data}
\label{DAMEX-2D}
\end{figure}

The purpose of this simulation is to provide an insight into the rationale of the algorithm in the bivariate case.
Normal data are simulated under a 2D logistic distribution with
asymmetric parameters (white and green points in
Fig.~\ref{DAMEX-2D}), while the abnormal ones are  uniformly
distributed. Thus, the `normal' extremes should be concentrated around
the axes, while the `abnormal' ones  could be anywhere. The
training set (white points) consists of normal observations. The
testing set consists of normal observations (white points) and 
abnormal ones (red points). 
Fig.~\ref{DAMEX-2D} represents the level sets of this scoring
function (inversed colors, the darker, the more abnormal) in both the transformed and the non-transformed input space.

\subsection{Recovering the support of the dependence structure}
In this section, we simulate data whose asymptotic behavior corresponds to some exponent measure $\mu$. This measure is chosen such that it concentrates on $K$ chosen cones. 
Experiments illustrate in this case how many data is needed to recover properly the $K$ sub-cones (namely the dependence structure) depending on its complexity. If the dependence structure spreads on a high number $K$ of sub-cones, then a high number of data will be required.

Datasets of size $50000$ (resp. $150000$) are  generated in $\mathbb{R}^{10}$ according to a popular multivariate extreme value
model, introduced by \cite{Tawn90},  namely a multivariate asymmetric
logistic distribution ($G_{log}$). 
The data have the following features: (i) They resemble `real life'
data, that is, the $X_i^j$'s are non
zero  and the transformed $\hat V_i$'s belong to the interior cone
$\mathcal{C}_{\{1,\ldots,d\}}$ (ii) The associated (asymptotic) exponent measure concentrates on
 $K$ disjoint cones $\{\mathcal{C}_{\alpha_m} , 1\le m\le K\}$.  
 For the sake of reproducibility, 
 $ G_{log}(\mb x) = \exp\{ - \sum_{m = 1}^K \left(\sum_{j \in \alpha_m}
     (|A(j)|x_j)^{ - 1/{w_{\alpha_m}}}\right)^{w_{\alpha_m}} \}, $
 where $|A(j)|$ is the cardinal of the set $\{\alpha\in D: j \in
 \alpha\}$ and where $w_{\alpha_m} = 0.1$ is a dependence parameter
 (strong dependence). 
The data are simulated using  Algorithm 2.2 in \cite{Stephenson2003}.
The subset of sub-cones $D$ with non-zero $\mu$-mass is randomly chosen (for each
fixed number of sub-cones $K$) and the purpose is to recover $D$ by Algorithm~\ref{DAMEX-algo}.
  For each $K$, $100$ experiments
are made and we consider  the  number of `errors', that is,      the number of
non-recovered or false-discovered sub-cones. Table~\ref{table:logevd} shows the averaged
numbers of errors  among the $100$ experiments. 
\begin{table}[h]
\centering
\begin{tabular}{|c|cc|}
  \hline
  $\#$ sub-cones $K$ & Aver. $\#$ errors  & Aver. $\#$ errors\\
     & (n=5e4)& (n=15e4) \\
\hline
3    &  0.07  & 0.01 \\
5    &  0.00  & 0.01 \\ 
10   &  0.01  & 0.06 \\
15   &  0.09  & 0.02 \\
20   &  0.39  & 0.14 \\
25   &  1.12  & 0.39 \\
30   &  1.82  & 0.98 \\
35   &  3.59  & 1.85 \\
40   &  6.59  & 3.14 \\
45   &  8.06  & 5.23 \\
50   &  11.21 & 7.87 \\
  \hline
\end{tabular}
\caption{Support recovering on simulated data}
\label{table:logevd}
\end{table}

The results are very promising in situations where the number of sub-cones is
moderate \emph{w.r.t.} the number of observations. 
Indeed, when the total number of sub-cones in the dependence structure is
`too large' (relatively to the number of observations),
some sub-cones are under-represented and become `too weak' to resist the thresholding (Step 4 in
Algorithm~\ref{DAMEX-algo}).  Handling  complex dependence
structures without a confortable number of observations thus requires a careful choice of the  thresholding level
$\mu_{\min}$, for instance by cross-validation.

\section{Real-world data sets}
\label{sec:experiments-real}
\subsection{Sparse structure of extremes  (wave data)}
Our goal is here to verify that the two expected phenomena mentioned
in the introduction, \textbf{1-}~sparse dependence structure of extremes (small number
of sub-cones with non zero mass), \textbf{2-}~low dimension of the
sub-cones with non-zero mass,  do occur with real data. 
\begin{figure}[!ht]
\centering
\includegraphics[scale=0.22]{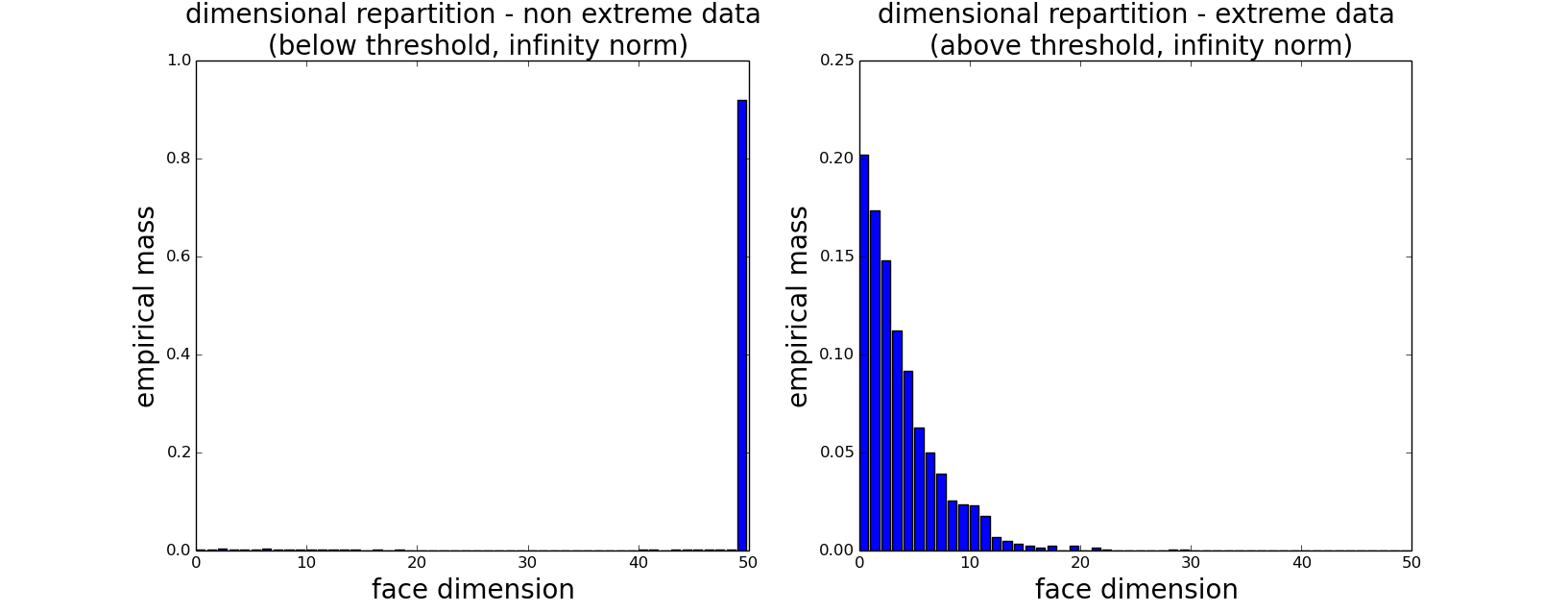}
\caption{sub-cone  dimensions of wave data}
\label{fig:wavedata-dim}
\end{figure}

We consider wave
directions data provided by Shell, which consist of $58585$
measurements  $D_i$, $i\le 58595$ of wave directions between $0^{\circ}$ and $360^{\circ}$ at $50$ different
locations (buoys in North sea). The dimension is thus $50$. 
The angle $90^{\circ}$ being fairly
rare, we work with data obtained as $X_i^j = 1/(10^{-10} + |90-
D_i^j|)$, where $D_i^j$ is the wave direction at buoy $j$, time $i$. Thus,
$D_i^j$'s close to $90$ correspond to  extreme $X_i^j$'s.
Results in
Table~\ref{fig:wavedata-nb-faces} ($\mu_{total}$ denotes the total probability mass of $\mu$)
show that, 
the 
number of  sub-cones $\mathcal{C}_\alpha$ identified by Algorithm~\ref{DAMEX-algo}
is indeed small compared to the total number of sub-cones ($2^{50}$-1)
(Phenomenon \textbf{1}). Extreme data are essentially concentrated in $18$ sub-cones.
Further, the dimension of those sub-cones is essentially moderate
(Phenomenon \textbf{2}):
respectively $93\%$, $98.6\%$ and  $99.6\%$
of the mass is affected to  sub-cones of dimension no greater  than $10$,
$15$ and $20$ respectively 
(to be compared with $d=50$).  Histograms displaying the mass repartition produced by Algorithm~\ref{DAMEX-algo} are given in Fig.~\ref{fig:wavedata-dim}. 
\begin{table}[!ht]
\centering
\begin{tabular}{|l|cc|}
\hline

~ & non-extreme & extreme \\
~ & data        &  data   \\
\hline
\footnotesize{ $\#$ of sub-cones   with positive  }       &      &     \\
\footnotesize{ mass   ($\mu_{\min}/\mu_{total} = 0$)}      & 3413 & 858 \\
\hline
\footnotesize{ ditto after  thresholding }  &   &    \\
\footnotesize{ ($\mu_{\min}/\mu_{total} = 0.002$)}                & 2 & 64 \\
\hline
\footnotesize{ ditto after  thresholding }   &   &    \\
\footnotesize{($\mu_{\min}/\mu_{total} = 0.005$)}  & 1 & 18 \\

\hline
\end{tabular}
\caption{Total number of sub-cones of wave data}
\label{fig:wavedata-nb-faces}
\end{table}


\subsection{Anomaly Detection}
\begin{figure}[!ht]
\centering
\includegraphics[scale=0.22]{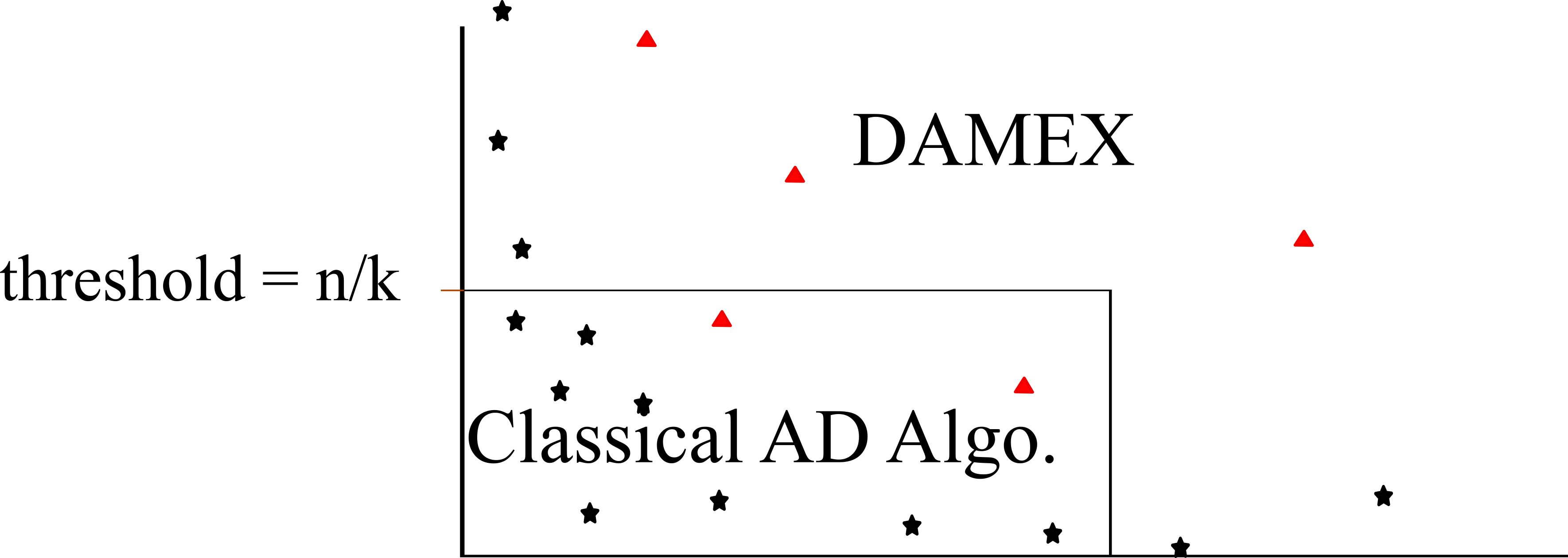}
\caption{Combination of any AD algorithm with DAMEX}
\label{fig:combination}
\end{figure}
 The main purpose of Algorithm~\ref{DAMEX-algo} is to deal with
extreme data. In this section we show that it may be combined with a
standard AD algorithm  to handle extreme \emph{and} non-extreme data,
improving the global performance of the chosen standard algorithm. 
This can be done as
illustrated in Fig.~\ref{fig:combination} by splitting the input
space between an extreme region and a non-extreme one, then applying
Algorithm~\ref{DAMEX-algo} to the extreme region, while the
non-extreme one is processed with the standard
algorithm.

One standard AD algorithm is the Isolation Forest (iForest)
algorithm, which we  chose 
in view of its
established 
high performances (\cite{Liu2008}). Our aim is to compare the results obtained with the
combined method
 `iForest + DAMEX' above described, to those obtained with iForest
 alone on the whole input space. 
\begin{table}[!ht]
\centering
\begin{tabular}{|l|cc|}
  \hline
  ~           & number of samples  & number of features \\
  shuttle     & 85849              & 9                  \\
  forestcover & 286048             & 54                 \\
  SA          & 976158             & 41                 \\
  SF          & 699691             & 4                  \\
  http        & 619052             & 3                  \\
  smtp        & 95373              & 3                  \\
  \hline
\end{tabular}
\caption{Datasets characteristics}
\label{table:data}
\end{table}

Six  reference datasets in AD are considered:
 \emph{shuttle}, \emph{forestcover}, \emph{http}, \emph{smtp},
 \emph{SF} and \emph{SA}. The experiments are performed in a
 semi-supervised framework (the training set consists of normal data). In a
 non-supervised framework (training set including abnormal data), the
 improvements brought by the use of DAMEX are less significant, but the
 precision score is still increased 
 when the recall is high (high rate of true positives), inducing more vertical ROC curves near the origin.


The \emph{shuttle} dataset is 
available in the UCI repository \cite{Lichman2013}. 
We use instances from all different classes but class $4$,  
which yields an anomaly ratio (class 1) of $7.15\%$. %
%
In the \emph{forestcover} data, also available at UCI
repository (\cite{Lichman2013}), the normal data are the  instances
from class~$2$ while instances from class $4$ are anomalies,
which yields an anomaly ratio of $0.9\%$. 
The last four datasets belong to the KDD Cup '99 dataset
(\cite{KDD99}, \cite{Tavallaee2009}), which consist of a wide variety
of hand-injected  attacks (anomalies) in a closed network (normal
background). Since the original demonstrative purpose of the dataset concerns
supervised AD, the anomaly rate is very high ($80\%$). 
We thus transform the KDD data to obtain smaller anomaly rates. For
datasets \emph{SF}, \emph{http} and \emph{smtp}, we proceed as
described in \cite{Yamanishi2000}: \emph{SF} is obtained by picking up the
data with positive logged-in attribute,
and 
focusing on the intrusion attack, which gives 
$0.3\%$
of anomalies.  The two datasets \emph{http} and \emph{smtp} are two subsets of \emph{SF}
corresponding to a third feature equal to 'http' (resp. to 'smtp').
Finally, the \emph{SA} dataset  is obtained as in \cite{Eskin2002} by 
selecting all the normal data, together with a small proportion
($1\%$) of anomalies. 

Table~\ref{table:data} summarizes the characteristics of these
datasets. For each of them, 20 experiments on random training and testing datasets are performed, yielding averaged ROC and Precision-Recall curves whose AUC are presented in Table~\ref{table:results-dbad+iforest}.
The parameter $\mu_{\min}$ is fixed to $\mu_{total}/ (\#$charged sub-cones), the averaged mass of the non-empty sub-cones.
\begin{table}[!ht]
\centering
\begin{tabular}{|l|cc|cc|}
  \hline
Dataset & \multicolumn{2}{c|}{iForest only} & \multicolumn{2}{c|}{iForest + DAMEX} \\
  \hline
~            & ROC          &  PR        & ROC        & PR      \\
shuttle      & 0.996        & 0.974      &$\mb{0.997}$&$\mb{0.987}$\\
forestcov.   & 0.964        & 0.193      &$\mb{0.976}$&$\mb{0.363}$\\
http         & 0.993        & 0.185      &$\mb{0.999}$&$\mb{0.500}$\\
smtp         & $\mb{0.900}$ &$\mb{0.004}$&0.898       &0.003       \\
SF           & 0.941        & 0.041      &$\mb{0.980}$&$\mb{0.694}$\\
SA           & 0.990        & 0.387      &$\mb{0.999}$&$\mb{0.892}$\\
  \hline
\end{tabular}
\caption{Results in terms of AUC}
\label{table:results-dbad+iforest}
\end{table}

The parameters $(k,\epsilon)$ are chosen according to remarks \ref{rk_param_interpretation} and \ref{rk_param_choice}. The stability \textit{w.r.t.} $k$ (resp. $\epsilon$) is investigated over the range [$n^{1/4}, n^{2/3}$] (resp. $[0.0001, 0.1]$). This yields parameters $( k, \epsilon) = (n^{1/3}, 0.0001)$ for \emph{SA} and \emph{forestcover}, and $(k, \epsilon) = (n^{1/2}, 0.01)$ for \emph{shuttle}. As the datasets \emph{http}, \emph{smtp} and \emph{SF} do not have enough features to consider the stability, we choose the (standard) parameters $(k, \epsilon) = (n^{1/2}, 0.01)$.
DAMEX significantly improves the precision
for each dataset, excepting for \emph{smtp}. 

\begin{figure}[!ht]
  \centering
  \includegraphics[width = 0.50 \textwidth]{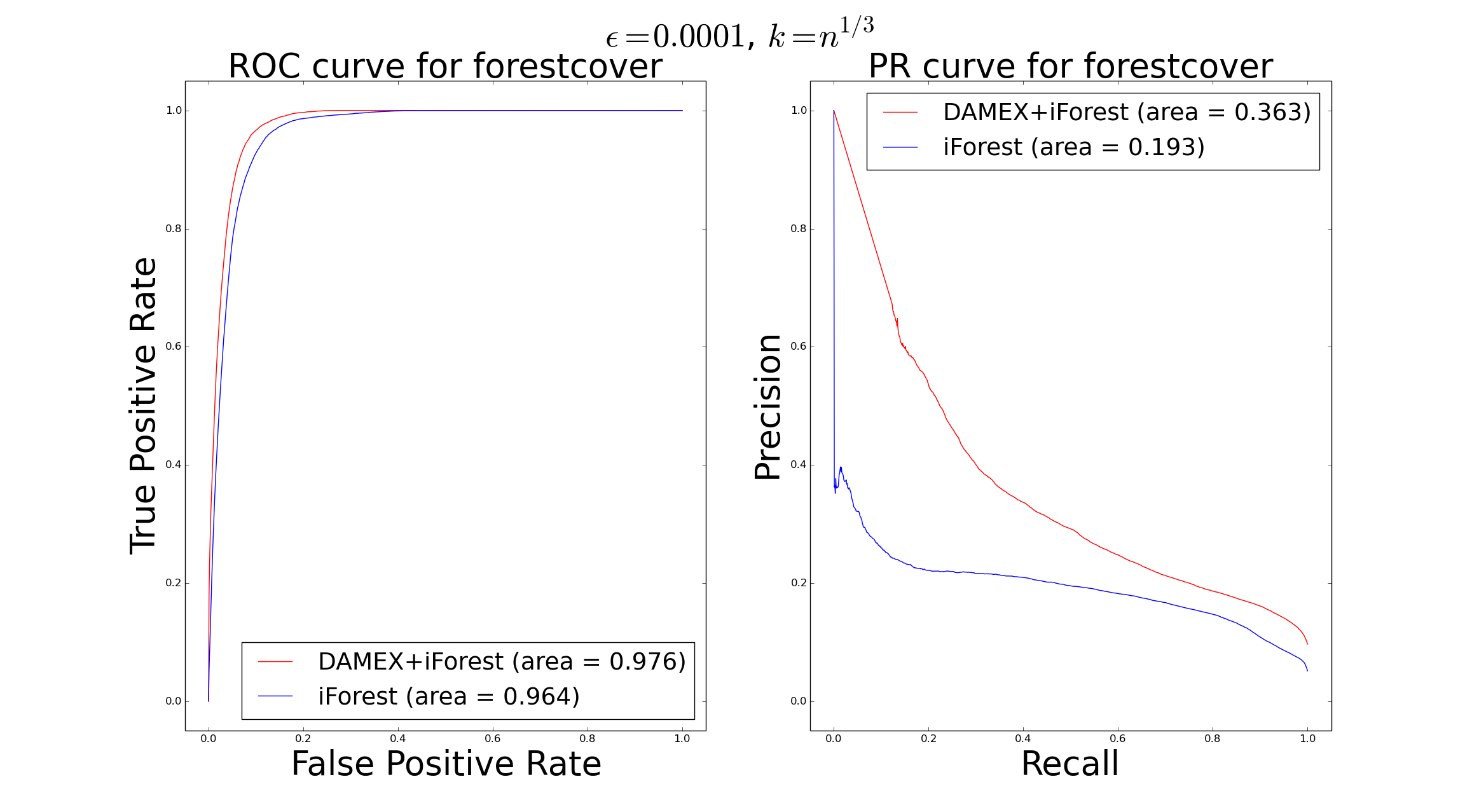}
  \caption{ROC and PR curve on forestcover dataset}
  \label{forestcover}
\end{figure}

\begin{figure}[!ht]
  \centering
  \includegraphics[width = 0.50 \textwidth]{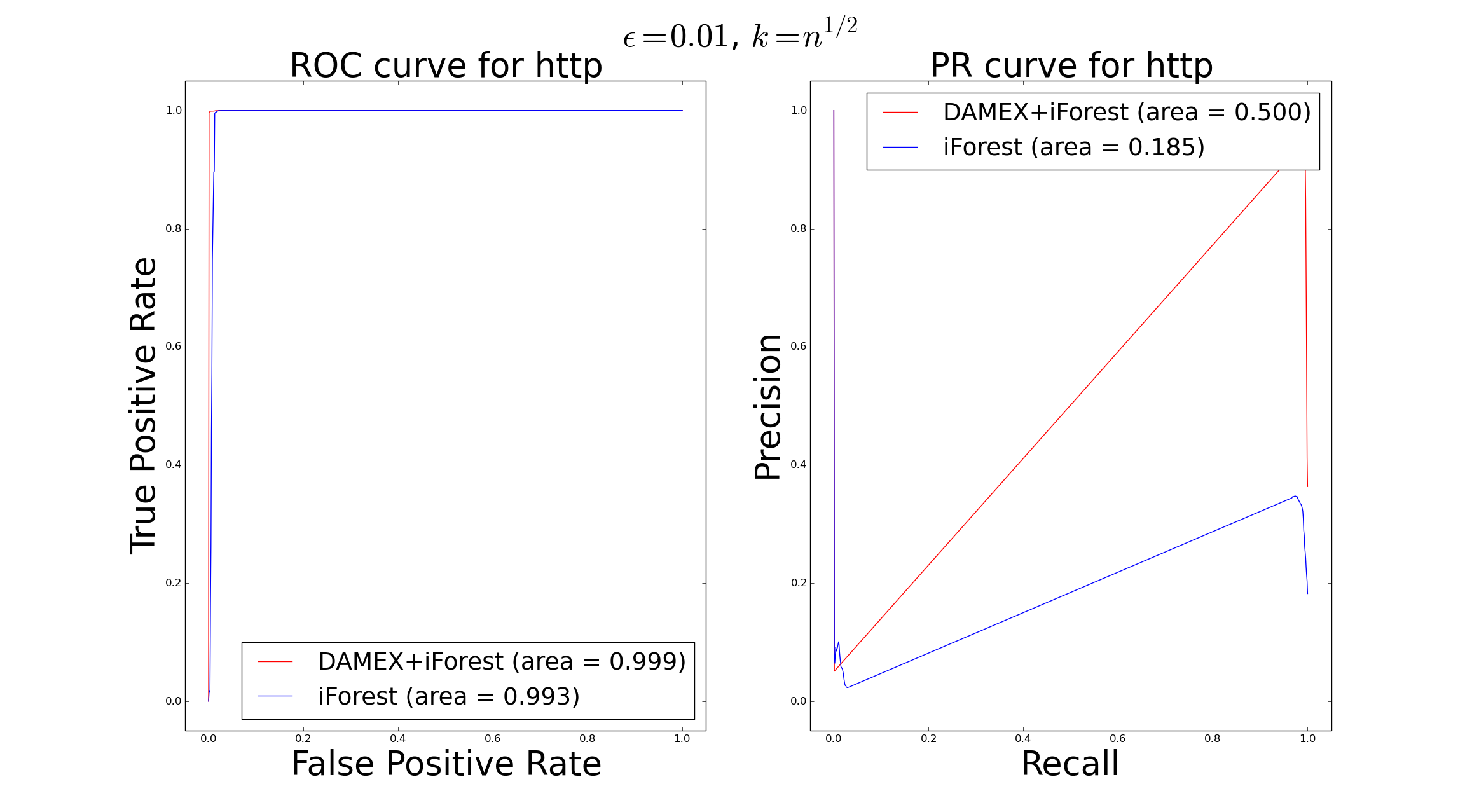}
  \caption{ROC and PR curve on http dataset}
  \label{http}
\end{figure}

\begin{figure}[!ht]
  \centering
  \includegraphics[width = 0.50 \textwidth]{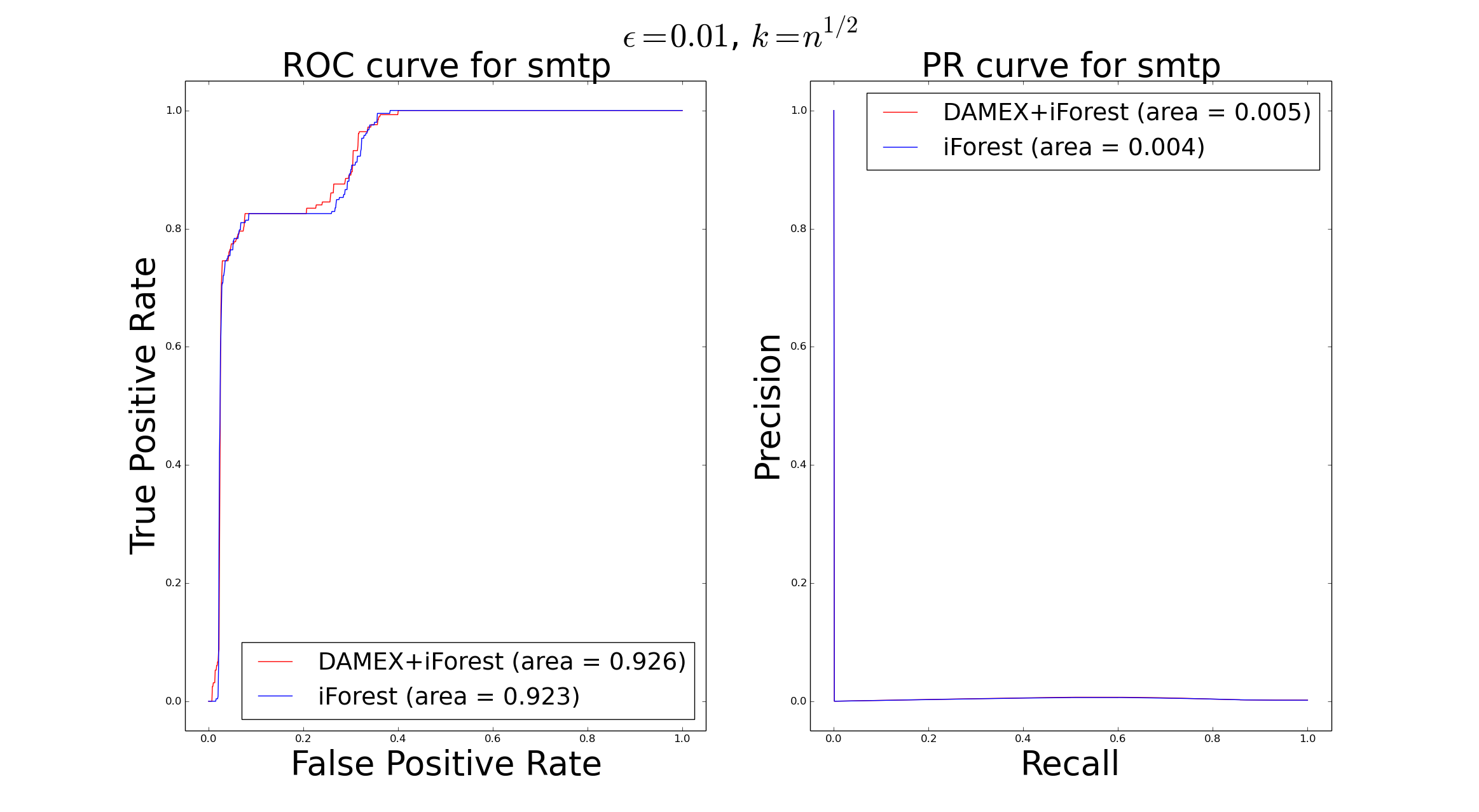}
  \caption{ROC and PR curve on smtp dataset}
  \label{smtp}
\end{figure}
\begin{figure}[!ht]
  \centering
  \includegraphics[width = 0.50 \textwidth]{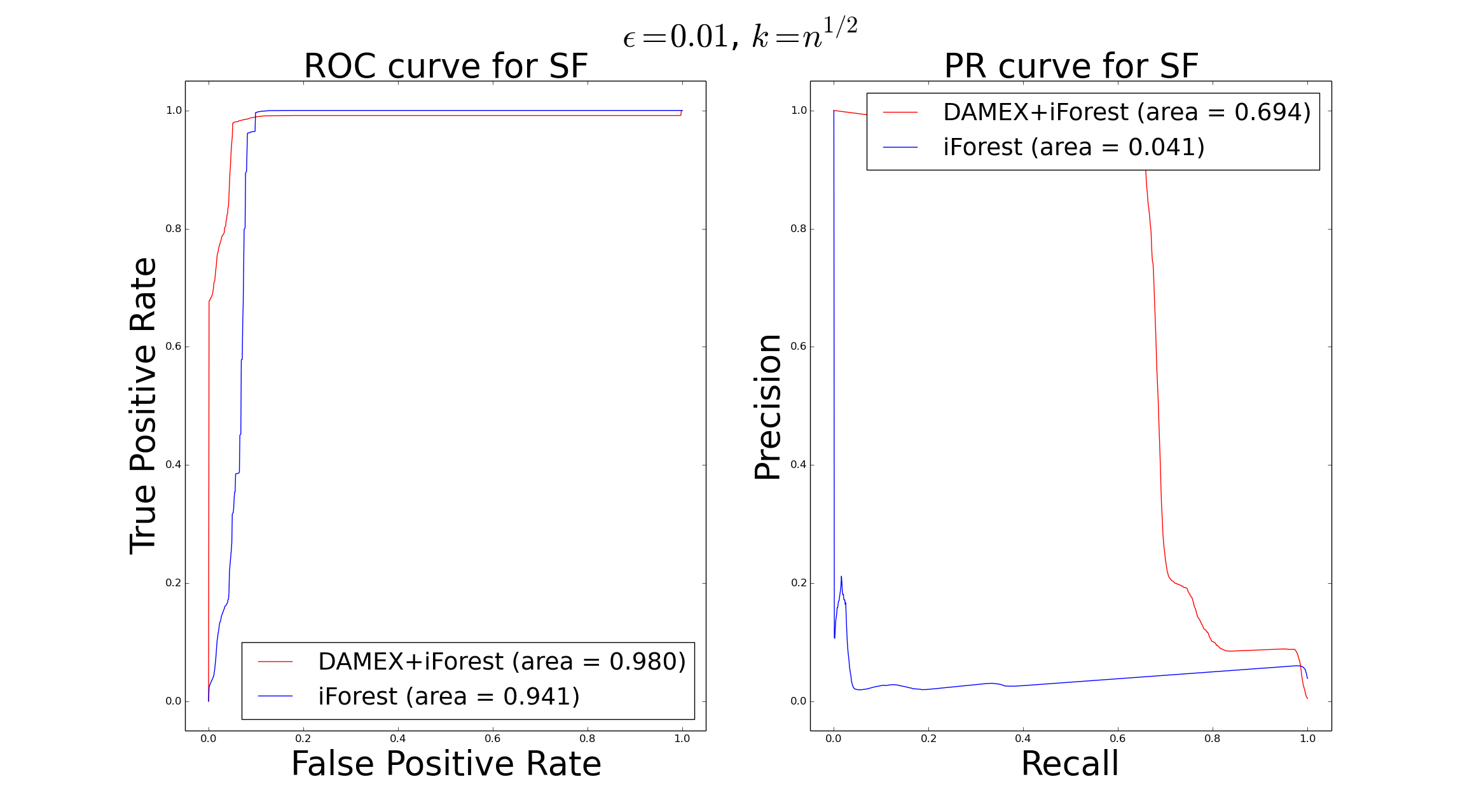}
  \caption{ROC and PR curve on SF dataset}
  \label{SF}
\end{figure}

In terms of AUC of the ROC curve, one observes slight or negligible improvements. Figures \ref{forestcover}, \ref{http}, \ref{smtp}, \ref{SF} represent averaged ROC curves and PR curves for \emph{forestcover}, \emph{http}, \emph{smtp} and \emph{SF}. The curves for the two other datasets are available in supplementary material. Excepting for the \emph{smtp} dataset, one observes highter slope at the origin of the ROC curve using DAMEX. 
It illustrates the fact that DAMEX is particularly adapted to situation where one has to work with a low false positive rate constrain.

Concerning the \emph{smtp} dataset, the algorithm seems to be unable to capture any extreme dependence structure, either because the latter is non-existent (no regularly varying tail), or because the convergence is too slow to appear in our relatively small dataset. 


\section{Conclusion}
The DAMEX algorithm allows to detect  anomalies occurring  in extreme
regions of a possibly large-dimensional input space by identifying
lower-dimensional subspaces around which normal extreme data concentrate. It is designed in accordance with 
well established results borrowed from multivariate Extreme Value Theory. 
 Various experiments on simulated data and real Anomaly Detection datasets demonstrate its ability
 to recover the support of the extremal dependence structure of the data, 
 thus improving the performance of standard Anomaly Detection algorithms. These
 results pave the way towards novel approaches in Machine Learning that
 take advantage of multivariate Extreme Value Theory tools for learning
 tasks involving features subject to extreme behaviors.

\subsubsection*{Acknowledgements}
Part of this work has been supported by the industrial chair `Machine Learning for Big Data' from Télécom ParisTech.
\clearpage
\subsubsection*{References}
\renewcommand\refname{\vskip -1cm} 
\bibliographystyle{plain}
\bibliography{mvextrem}

\end{document}